\documentclass[11pt]{article}

\usepackage[margin=1in]{geometry}

\usepackage[T1]{fontenc}
\usepackage[utf8]{inputenc}
\usepackage{lmodern}             
\usepackage{microtype}  
\usepackage{textcomp}

\usepackage{booktabs}             
\usepackage{longtable}            
\usepackage{array}                

\usepackage{graphicx}
\usepackage{float}
\usepackage{placeins}             
\usepackage[font=scriptsize,labelfont=bf]{caption}  

\usepackage{fancyvrb}
\fvset{fontsize=\scriptsize}      
\RecustomVerbatimEnvironment{verbatim}{Verbatim}{}

\usepackage{amsmath,amssymb}

\usepackage{hyperref}
\hypersetup{
    colorlinks=true,
    linkcolor=blue,
    citecolor=blue,
    urlcolor=blue
}
\usepackage{url}

\usepackage{CJKutf8}

\makeatletter
\def\maxwidth{\ifdim\Gin@nat@width>\linewidth\linewidth\else\Gin@nat@width\fi}
\def\maxheight{\ifdim\Gin@nat@height>\textheight\textheight\else\Gin@nat@height\fi}
\makeatother
\setkeys{Gin}{width=\maxwidth,height=\maxheight,keepaspectratio}

\providecommand{\tightlist}{}

\title{\textbf{JP-TL-Bench: Anchored Pairwise LLM Evaluation for Bidirectional Japanese-English Translation}}
\author{Leonard Lin, Adam Lensenmayer\\ Shisa.AI}
\date{}

\begin{document}
\begin{CJK*}{UTF8}{min}

\maketitle

\begin{abstract}
We introduce \textbf{JP-TL-Bench}, a lightweight, open benchmark designed to guide the \textbf{iterative development} of Japanese\(\leftrightarrow\)English translation systems. In this context, the challenge is often ``which of these two good translations is better?'' rather than ``is this translation acceptable?'' This distinction matters for Japanese\(\leftrightarrow\)English, where subtle choices in politeness, implicature, ellipsis, and register strongly affect perceived naturalness. JP-TL-Bench uses a protocol built to make LLM judging both \textbf{reliable} and \textbf{affordable}: it evaluates a candidate model via \textbf{reference-free, pairwise LLM comparisons} against a \textbf{fixed, versioned anchor set}. Pairwise results are aggregated with a \textbf{Bradley--Terry model} \cite{bradley1952rank} and reported as win rates plus a normalized \textbf{0--10 ``LT'' score} derived from a logistic transform of fitted log-strengths. Because each candidate is scored against the same frozen anchor set, scores are \textbf{structurally stable} given the same base set, judge, and aggregation code.
\end{abstract}

\section{1. Introduction}\label{introduction}
\FloatBarrier

During development of the Shisa V2 bilingual Japanese-English models \cite{shisav2,shisav2405b,shisav21}, we found existing evaluation approaches inadequate for answering a practical question: ``which of these two good translations is better?'' Large language models (LLMs) have dramatically improved machine translation (MT) quality, but much of MT evaluation has not kept pace. Suites like llm-jp-eval \cite{llmjpeval} provide valuable COMET-based validation, but these benchmarks are largely saturated: scores cluster tightly and often fail to separate strong translations.

The MT community relies heavily on reference-based metrics such as BLEU \cite{papineni-etal-2002-bleu}, chrF \cite{popovic-2015-chrf}, and learned metrics such as COMET \cite{rei-etal-2020-comet}. These are useful for broad validation, but recent work shows they can mischaracterize quality at the top-end and are not designed to provide high-resolution signals for outputs that are already near-fluent \cite{agrawal-etal-2024-automatic-metrics,zouhar2024cometpitfalls,reiter2018structured}. Modern frontier LLMs can recognize subtle nuance, making them attractive as reference-free judges---however LLM-as-a-judge \cite{zheng2024judging} can be unreliable without careful protocol design \cite{llmjudge_survey}.

\subsection{1.1 Overview}\label{overview}

We present \textbf{JP-TL-Bench}, a benchmark and protocol aimed at the specific use case of \textbf{iterating on Japanese\(\leftrightarrow\)English translation quality} when absolute metrics become hard to interpret. Pairwise LLM-as-a-judge comparisons offer strong discrimination and reliability; the key idea here is to compare against a frozen anchor set rather than all-pairs, which yields stable absolute scores at fixed O(N) cost. Large-scale preference leaderboards such as Chatbot Arena \cite{chiang2024chatbot} have proven the utility of pairwise comparisons, but their Elo-based rankings are order-dependent, produce floating scores that drift as the pool changes, and scale O(N\textsuperscript{2}) in cost. By using a fixed anchor set, JP-TL-Bench avoids all three issues.

The benchmark contains 70 translation items spanning \textbf{EN\(\to\)JA and JA\(\to\)EN} and \textbf{Easy/Hard} difficulty tiers. A single evaluation run requires approximately \textbf{70 \(\times\) 20 = 1,400} pairwise judgments per candidate model (linear in the number of evaluated models), making the benchmark practical for iterative fine-tuning. We additionally describe two complementary baselines often used in practice---\textbf{llm-jp-eval} \cite{llmjpeval} (COMET-centered suite) and \textbf{rubric-based LLM judging} (as used in our LiquidAI hackathon workflow)---and provide prompt/rubric templates in the appendix, connecting to rubric-trained judges and feedback datasets such as Prometheus \cite{kim2024prometheus} and UltraFeedback \cite{cui2024ultrafeedback}.

\subsection{1.2 Contributions}\label{contributions}

\begin{itemize}
\tightlist
\item
  \textbf{Anchored pairwise protocol} for Japanese\(\leftrightarrow\)English translation evaluation: candidate models are compared against a fixed, versioned anchor set rather than a floating pool.
\item
  \textbf{Reference-free judging} with a transparent prompt and deterministic decoding; the benchmark operates without reference translations.
\item
  \textbf{Score aggregation and reporting} via Bradley--Terry \cite{bradley1952rank} with a normalized 0--10 LT score for interpretability and comparability under fixed conditions.
\item
  \textbf{Curated anchor set}: Base Set v1.0 was selected from hundreds of models to provide evenly-spaced win rates across a broad quality range (as of mid-2025).
\item
  \textbf{Versioned comparison pools}: Anchor sets use semantic versioning to allow fixes and improvements while preserving comparability across evaluations.
\item
  \textbf{Open-source implementation}: JP-TL-Bench has been used in the development of several Shisa.AI model releases and is available under Apache 2.0 at \url{https://github.com/shisa-ai/jp-tl-bench.}
\item
  \textbf{Practical integration guidance}: How JP-TL-Bench complements llm-jp-eval \cite{llmjpeval} and rubric-based LLM judging (Appendix C).
\end{itemize}

\vspace{0.5em}

\section{2. Background and Related Work}\label{background-and-related-work}
\FloatBarrier

\subsection{2.1 Reference-based MT metrics and validity concerns}\label{reference-based-mt-metrics-and-validity-concerns}

BLEU \cite{papineni-etal-2002-bleu} and chrF \cite{popovic-2015-chrf} remain widely used despite known limitations (surface overlap, sensitivity to tokenization and valid paraphrase)---indeed, the WMT22 Metrics Shared Task concluded that overlap metrics correlate poorly with human ratings and recommends moving beyond them \cite{freitag-etal-2022-results}. Learned metrics such as COMET \cite{rei-etal-2020-comet} correlate well with human judgments on WMT-style settings, but recent analyses raise concerns about metric behavior in high-quality regimes and evaluation pitfalls \cite{agrawal-etal-2024-automatic-metrics,zouhar2024cometpitfalls}. More broadly, the validity of BLEU as an evaluation instrument has been critically reviewed \cite{reiter2018structured}.

The COMET family also includes reference-free quality estimation variants such as CometKiwi \cite{rei2022cometkiwi} and more transparent variants such as xCOMET \cite{guerreiro2023xcomet}; in Section 5.2 we compare COMET-scoring to our JP-TL-Bench LLM Judge results.

\subsection{2.2 Human evaluation frameworks}\label{human-evaluation-frameworks}

Human evaluation remains the gold standard. MQM provides a structured framework for fine-grained error annotation \cite{lommel2014mqm}, and direct assessment techniques establish continuous human scoring protocols \cite{graham-etal-2013-continuous}. However, expert evaluation is costly and difficult to scale; large-scale evidence highlights the complexity of obtaining reliable human judgments \cite{freitag2021experts}.

\subsection{2.3 LLM-based MT evaluation}\label{llm-based-mt-evaluation}

LLM-based evaluation has emerged as a scalable alternative. Kocmi and Federmann show that strong LLMs can act as high-performing translation evaluators \cite{kocmi2023llm_eval_mt}, and GEMBA-MQM extends this idea with MQM-style error spans \cite{kocmi-federmann-2023-gemba-mqm}. BatchGEMBA explores token-efficient judging via batching and prompt compression \cite{larionov2025batchgemba}.

\subsection{2.4 LLM-as-a-judge reliability, bias, and rubrics}\label{llm-as-a-judge-reliability-bias-and-rubrics}

LLM-as-a-judge has been studied in general evaluation settings \cite{zheng2024judging,schroeder2024trust} and surveyed \cite{llmjudge_survey}. Judge bias and unfairness concerns are documented \cite{wang2023large}, and position bias is a known failure mode in comparative evaluation settings \cite{ko2020position}. A parallel line of work trains or distills specialized judges using rubrics and feedback datasets, e.g., Prometheus \cite{kim2024prometheus} and UltraFeedback \cite{cui2024ultrafeedback}.

\subsection{2.5 Pairwise preference leaderboards and ranking models}\label{pairwise-preference-leaderboards-and-ranking-models}

Pairwise preference evaluation at scale is popularized by Chatbot Arena \cite{chiang2024chatbot}. Preference aggregation often uses Bradley--Terry-style models \cite{bradley1952rank} or Elo-style updates \cite{elo1978rating}. A key distinction for translation benchmarking is \textbf{score stability}: floating-pool leaderboards can drift as the pool changes, while anchored benchmarks trade some flexibility for comparability.

FiRE \cite{fire2026} proposes fine-grained, reference-free ranking evaluation for MT, making it a close methodological neighbor; JP-TL-Bench differs by explicitly anchoring scores to a frozen base set snapshot to support stable iteration.

\subsection{2.6 Japanese MT suites and benchmarks}\label{japanese-mt-suites-and-benchmarks}

FLORES/NLLB provide multilingual MT benchmarks including Japanese \cite{goyal2022flores101,nllb2022}. The llm-jp-eval project \cite{llmjpeval} provides a Japanese evaluation suite that includes MT tasks and standardized scoring (COMET-centered), and is commonly used as a reference implementation for MT validation. Parallel data resources such as JParaCrawl \cite{morishita-etal-2020-jparacrawl} support training and evaluation, but do not solve the fine-grained discrimination problem by themselves.

Japanese\(\leftrightarrow\)English translation poses challenges that sentence-level evaluations often miss. Japanese is a pro-drop language where subjects and objects are frequently omitted (zero pronouns), requiring inference from discourse context \cite{wang-etal-2018-prodrop}. JA\(\to\)EN systems often resolve these incorrectly, producing misgendered pronouns or swapped thematic roles---errors that sentence-level metrics like BLEU fail to capture \cite{shimazu-etal-2020-zero-pronoun}. Japanese is also highly register-sensitive: honorific speech (keigo) encodes respect, formality, and social distance through verb morphology. Since English lacks grammaticalized honorifics, EN\(\to\)JA systems frequently produce inappropriate formality levels---either overly casual output or misapplied keigo---without explicit control mechanisms \cite{feely-etal-2019-keigo}.

\subsection{2.7 Summary comparison}\label{summary-comparison}

\begin{table}[htbp]
\centering
\small
\caption{MT Evaluation Approaches}
\begin{tabular}{@{}lllll@{}}
\toprule
Approach & Method & Judge & Cost & Limitations \\
\midrule
BLEU/chrF & N-gram overlap with reference & Algorithmic & Low & Penalizes valid paraphrases \\
COMET & Neural embedding similarity & Neural model & Low & Saturates at high quality \\
MQM Human & Expert error annotation & Human & High & Expensive, slow \\
GEMBA & LLM absolute scoring (0-100) & LLM & Medium & Score compression at top \\
Chatbot Arena & Pairwise + floating Elo & Human & High & Order-dependent, score drift, O(N\textsuperscript{2}) \\
\textbf{JP-TL-Bench} & Pairwise + fixed anchors & LLM & Low & Base set/judge dependent \\
\bottomrule
\end{tabular}
\end{table}

\vspace{0.5em}

\section{3. JP-TL-Bench Benchmark Design}\label{jp-tl-bench-benchmark-design}
\FloatBarrier

\subsection{3.1 Task and items}\label{task-and-items}

JP-TL-Bench contains 70 translation items designed to stress Japanese-specific phenomena (register/keigo, ambiguity resolution, cultural adaptation, technical terminology). Items are split across:

\begin{itemize}
\tightlist
\item
  \textbf{Direction}: EN\(\to\)JA (34 items) and JA\(\to\)EN (36 items)
\item
  \textbf{Difficulty}: Easy (30 items) vs Hard (40 items)
\end{itemize}

The breakdown by slice: EN\(\to\)JA Easy (15), EN\(\to\)JA Hard (19), JA\(\to\)EN Easy (15), JA\(\to\)EN Hard (21).

The test corpus was constructed by the authors after reviewing existing MT corpora and observing real-world failure modes from prior Shisa model deployments. Item selection was guided by one of the authors, who brings over 10 years of professional experience as a Japanese translator, focusing on phenomena that distinguish adequate from excellent translations. Sample items are provided in Appendix A.

The item set is intentionally \textbf{small enough} to run frequently during model iteration, and \textbf{targeted enough} to separate strong systems where corpus-level metrics often provide limited spread.

\subsection{3.2 Anchor set (Base Set v1.0)}\label{anchor-set-base-set-v1.0}

The benchmark's core stability mechanism is a frozen \textbf{anchor set} of 20 models. The set includes both strong and weak systems to provide a wide dynamic range and to avoid over-clustering at the top end.

\textbf{Construction process}: Base Set v1.0 was derived from an initial v0.9 snapshot that accumulated translation outputs and pairwise results from nearly 200 models. From this pool, we manually selected 20 anchors to achieve approximately even win-rate spacing (\textasciitilde5\% intervals from \textasciitilde2\% to \textasciitilde96\%), balancing both EN\(\to\)JA and JA\(\to\)EN performance. This is sometimes challenging because directional performance can be highly asymmetric (e.g., a model strong at JA\(\to\)EN but weak at EN\(\to\)JA). The final selection prioritizes models that are reasonably balanced or represent distinct quality tiers in each direction.

The v1.0 manifest, translations, and scoring reports are available in the repository at \url{https://github.com/shisa-ai/jp-tl-bench} under \texttt{baseset/v1.0/}. Anchor win rates and LT scores below are taken from the published v1.0 report and correspond to the \textbf{gemini-2.5-flash} judge at \textbf{temperature 0}.

\begin{table}[htbp]
\centering
\small
\caption{Base Set v1.0 anchors (overall slice)}
\begin{tabular}{@{}rlrr@{}}
\toprule
\# & Model & Win Rate & LT \\
\midrule
1 & \texttt{google/gemini-2.5-pro} & 96.15\% & 9.94 \\
2 & \texttt{google/gemini-2.5-flash} & 92.93\% & 9.89 \\
3 & \texttt{Qwen/Qwen3-30B-A3B-Instruct-2507} & 84.37\% & 9.63 \\
4 & \texttt{shisa-ai/shisa-v2-llama3.1-405b} & 81.46\% & 9.49 \\
5 & \texttt{openai/gpt-4o} & 76.04\% & 9.12 \\
6 & \texttt{shisa-ai/shisa-v2-unphi4-14b} & 72.82\% & 8.83 \\
7 & \texttt{tokyotech-llm/Llama-3.1-Swallow-8B-Instruct-v0.5} & 62.14\% & 7.42 \\
8 & \texttt{nvidia/NVIDIA-Nemotron-Nano-12B-v2} & 59.94\% & 7.05 \\
9 & \texttt{meta-llama/Llama-3.3-70B-Instruct} & 58.05\% & 6.72 \\
10 & \texttt{microsoft/phi-4} & 49.80\% & 5.12 \\
11 & \texttt{cyberagent/Mistral-Nemo-Japanese-Instruct-2408} & 47.60\% & 4.67 \\
12 & \texttt{Qwen/Qwen3-4B} & 44.78\% & 4.10 \\
13 & \texttt{LiquidAI/LFM2-2.6B} & 43.83\% & 3.91 \\
14 & \texttt{meta-llama/Llama-3.1-8B-Instruct} & 38.81\% & 2.94 \\
15 & \texttt{microsoft/Phi-4-mini-instruct} & 24.98\% & 0.99 \\
16 & \texttt{augmxnt/shisa-7b-v1} & 21.44\% & 0.68 \\
17 & \texttt{meta-llama/Llama-3.2-3B-Instruct} & 19.24\% & 0.54 \\
18 & \texttt{Rakuten/RakutenAI-2.0-mini-instruct} & 14.23\% & 0.29 \\
19 & \texttt{LiquidAI/LFM2-350M} & 8.88\% & 0.14 \\
20 & \texttt{SakanaAI/TinySwallow-1.5B} & 2.51\% & 0.04 \\
\bottomrule
\end{tabular}
\end{table}

\textbf{Versioning}: Anchor sets follow semantic versioning. Patch versions (e.g., v1.0.1) may fix data errors or backfill missing judgments without changing the anchor model set; scores remain comparable. Minor versions (e.g., v1.1) may adjust anchor composition while maintaining rough calibration. Major versions (e.g., v2.0) indicate a new anchor pool where scores are not directly comparable to prior versions. This versioning contract allows researchers to cite specific snapshots while enabling the benchmark to evolve.

\vspace{0.5em}

\section{4. Evaluation Protocol}\label{evaluation-protocol}
\FloatBarrier

\subsection{4.1 Pair construction and A/B randomization}\label{pair-construction-and-ab-randomization}

For each item, the candidate model output is compared against each anchor model output, producing \textasciitilde1,400 pairs per candidate (70 \(\times\) 20). To mitigate position bias \cite{ko2020position}, the comparer \textbf{randomizes which side is ``Translation A'' vs ``Translation B'' once per pair using a fixed seed} (seed=42) for reproducibility; it does not require double-judging each pair in both orders.

\subsection{4.2 LLM judge prompt and decoding}\label{llm-judge-prompt-and-decoding}

JP-TL-Bench uses a fixed compare prompt (Appendix B) emphasizing eight dimensions (accuracy, naturalness, tone/register, etc.). Prior versions used a local LLM jury approach inspired by PoLL \cite{verga2024poll}, but as of Base Set v1.0, the default judge is \texttt{gemini-2.5-flash} with deterministic decoding (temperature 0). Deterministic decoding reduces run-to-run variance and makes stability primarily a property of the comparison set and aggregation.

\textbf{Evaluation cost and time}: With \textasciitilde1,400 judgments per candidate model and gemini-2.5-flash pricing at \$0.30/1M input tokens and \$2.50/1M output tokens (as of late 2025), a full evaluation run costs approximately \textbf{\textasciitilde\$7.00 USD} per candidate. Evaluations typically complete in \textbf{10--30 minutes} depending on model inference stack, output length, and judging concurrency. See Section 4.6 for detailed cost breakdown.

\subsection{4.3 Bradley--Terry aggregation}\label{bradleyterry-aggregation}

Pairwise outcomes are aggregated with a Bradley--Terry model \cite{bradley1952rank}. Let \(\theta_i\) be the fitted log-strength for model i. Then:

\[P(i \succ j) = \frac{\exp(\theta_i)}{\exp(\theta_i) + \exp(\theta_j)}.\]

The implementation uses maximum-likelihood estimation via \texttt{choix} \cite{choix}. Importantly, \textbf{each candidate is scored independently}: to score a candidate under a given Base Set version, we reuse the frozen anchor--anchor judgments from that Base Set and add the candidate--anchor judgments, then fit a Bradley--Terry model on this combined graph. We never mix judgments from different candidates in a single fit, so adding or removing other candidates cannot change an existing candidate's score.

\subsection{4.4 Reporting: win rate and LT score}\label{reporting-win-rate-and-lt-score}

JP-TL-Bench reports:

\begin{itemize}
\tightlist
\item
  \textbf{Win rate}: empirical wins / matches for the slice (overall, EN\(\to\)JA, JA\(\to\)EN, Easy, Hard).
\item
  \textbf{LT score (0--10)}: a logistic transform of centered log-strengths:
\end{itemize}

\[\mathrm{LT}_i = 10 \cdot \sigma(\theta_i - \overline{\theta}), \quad \sigma(x)=\frac{1}{1+e^{-x}}.\]

Here \(\overline{\theta}\) is a centering constant; in our implementation we take it to be the mean fitted log-strength for the slice over all models in that fit (the 20 anchors plus the candidate). The transform compresses extreme strengths while preserving ordering and improving interpretability on a bounded 0--10 scale.

\textbf{Note on aggregation}: Both Overall LT and Win Rate are computed directly over all matches. However, the Bradley-Terry model accounts for opponent strength---a win against a strong anchor contributes more to the fitted score than a win against a weak anchor---while Win Rate treats all wins equally. This can produce minor discrepancies between LT ranking and win-rate ranking.

\textbf{Handling empty outputs and judge refusals}: If a candidate model refuses to translate or produces empty output, the judge prompt instructs it to count this as a loss for that candidate (see Appendix B). However, if the \emph{judge itself} declines to evaluate a pair (e.g., due to safety filters on the source text), that judgment is excluded from aggregation rather than penalizing either side; this can slightly affect per-model match counts.

\subsection{4.5 Structural stability and comparability contract}\label{structural-stability-and-comparability-contract}

Elo-style systems \cite{elo1978rating} suffer from \textbf{score drift}: each model starts with an initial rating that adjusts based on match outcomes and opponent strength, so as new models enter the pool and shift the rating distribution, the meaning of a given score changes over time---a model rated 1200 today may not be comparable to one rated 1200 six months ago. Because JP-TL-Bench compares each candidate against a fixed anchor set rather than a floating pool, scores are structurally stable (order-independent) given:

\begin{enumerate}
\def\labelenumi{\arabic{enumi}.}
\tightlist
\item
  \textbf{Base Set version} (e.g., \texttt{baseset/v1.0})
\item
  \textbf{Judge model + prompt + decoding settings} (including temperature)
\item
  \textbf{Aggregation implementation/version}
\end{enumerate}

The fixed anchor set is the key insight: it combines the discriminative power of pairwise preference judgments \cite{zheng2024judging} with the temporal comparability of static benchmarks---something floating-pool leaderboards cannot provide.

\subsection{4.6 Complexity and cost}\label{complexity-and-cost}

Each candidate requires 70 prompts \(\times\) 20 anchors = \textbf{1,400 pairwise judgments}, scaling O(N) rather than O(N\textsuperscript{2}) for full round-robin. To estimate judge cost, we count tokens on the actual v1.0 judged Base Set (14,002 A/B judgments) using the exact comparison prompt (input: prompt template + formatted translations) and the judge's response (output: analysis). Across this data, mean token usage is 2,626 input and 1,567 output tokens per judgment.

Using gemini-2.5-flash pricing (\$0.30/M input, \$2.50/M output), this corresponds to:

\begin{table}[htbp]
\centering
\normalsize
\begin{tabular}{@{}lll@{}}
\toprule
Component & Tokens & Cost \\
\midrule
Input (2,626 $\times$ 1,400) & 3.68M & \textasciitilde\$1.10 \\
Output (1,567 $\times$ 1,400) & 2.19M & \textasciitilde\$5.48 \\
\textbf{Total per model} & 5.87M & \textbf{\textasciitilde\$6.59} \\
\bottomrule
\end{tabular}
\end{table}

As models are always compared against the same-sized limited comparison pool, the per-candidate judging cost remains roughly constant, making it feasible to run JP-TL-Bench during model development.

\subsection{4.7 Judgment quality}\label{judgment-quality}

During development, we compared LLM-as-a-judge results to native-speaker bilingual evaluations on a subset of items and found them comparable in both ratings and inter-rater agreement.

That said, while automated benchmarks make it practical to evaluate the hundreds of ablations generated during training at reasonable cost, we built multiple terminal-based (TUI) tools for JP-TL-Bench to allow convenient inspection and comparison of the judgments and outputs. A numeric score or win/loss record cannot replace examining actual model outputs and we encourage all researchers to do so (and all benchmark creators to build tools that make qualitative inspection convenient).

\vspace{0.5em}

\section{5. Snapshot Analysis and Benchmark Judgment}\label{snapshot-analysis-and-benchmark-judgment}
\FloatBarrier

\subsection{5.1 Dynamic range in the anchor set}\label{dynamic-range-in-the-anchor-set}

A key consideration for designing our Base Set v1.0 anchor set was giving it both the largest ``dynamic range'' (LT \(\approx\) 0.04 to 9.94) and relatively even spacing of Win-Loss ratios to allow for more even discrimination across the quality spectrum. This is important because poor set choice leads to unhelpful score clustering.

\begin{table}[htbp]
\centering
\scriptsize
\caption{Example slice scores (LT) from Base Set v1.0}
\begin{tabular}{@{}lrrrrrr@{}}
\toprule
Model & EN$\to$JA & EN$\to$JA Easy & EN$\to$JA Hard & JA$\to$EN & JA$\to$EN Easy & JA$\to$EN Hard \\
\midrule
gemini-2.5-pro & 9.97 & 9.95 & 9.99 & 9.89 & 9.79 & 9.99 \\
Llama-3.1-Swallow-8B-Instruct-v0.5 & 8.80 & 8.68 & 9.07 & 5.96 & 5.54 & 6.33 \\
LFM2-2.6B & 5.22 & 6.38 & 4.06 & 2.97 & 3.86 & 2.07 \\
Llama-3.1-8B-Instruct & 1.40 & 1.07 & 1.68 & 4.52 & 5.87 & 3.58 \\
\bottomrule
\end{tabular}
\end{table}

Two recurring patterns emerge:

\begin{itemize}
\tightlist
\item
  \textbf{Direction asymmetry}: Some models show much stronger performance in one direction. Llama-3.1-8B-Instruct scores 4.52 on JA\(\to\)EN but only 1.40 on EN\(\to\)JA---a 3.1 point gap. Llama-3.1-Swallow-8B shows the opposite pattern: 8.80 EN\(\to\)JA vs 5.96 JA\(\to\)EN (2.8 point gap), suggesting it was more heavily optimized for English-to-Japanese translation.
\item
  \textbf{Tier sensitivity}: Hard prompts tend to widen gaps among weaker models and can expose brittleness not visible on Easy prompts. LFM2-2.6B scores 6.38 on EN\(\to\)JA Easy but drops to 4.06 on EN\(\to\)JA Hard---a pattern common in smaller models that becomes less pronounced as models get stronger (compare to gemini-2.5-pro's 9.95/9.99 Easy/Hard consistency).
\end{itemize}

These patterns inform model selection for specific deployment scenarios and reveal quality dimensions invisible to aggregate metrics.

\begin{figure}[H]
\centering
\includegraphics[width=0.70\textwidth]{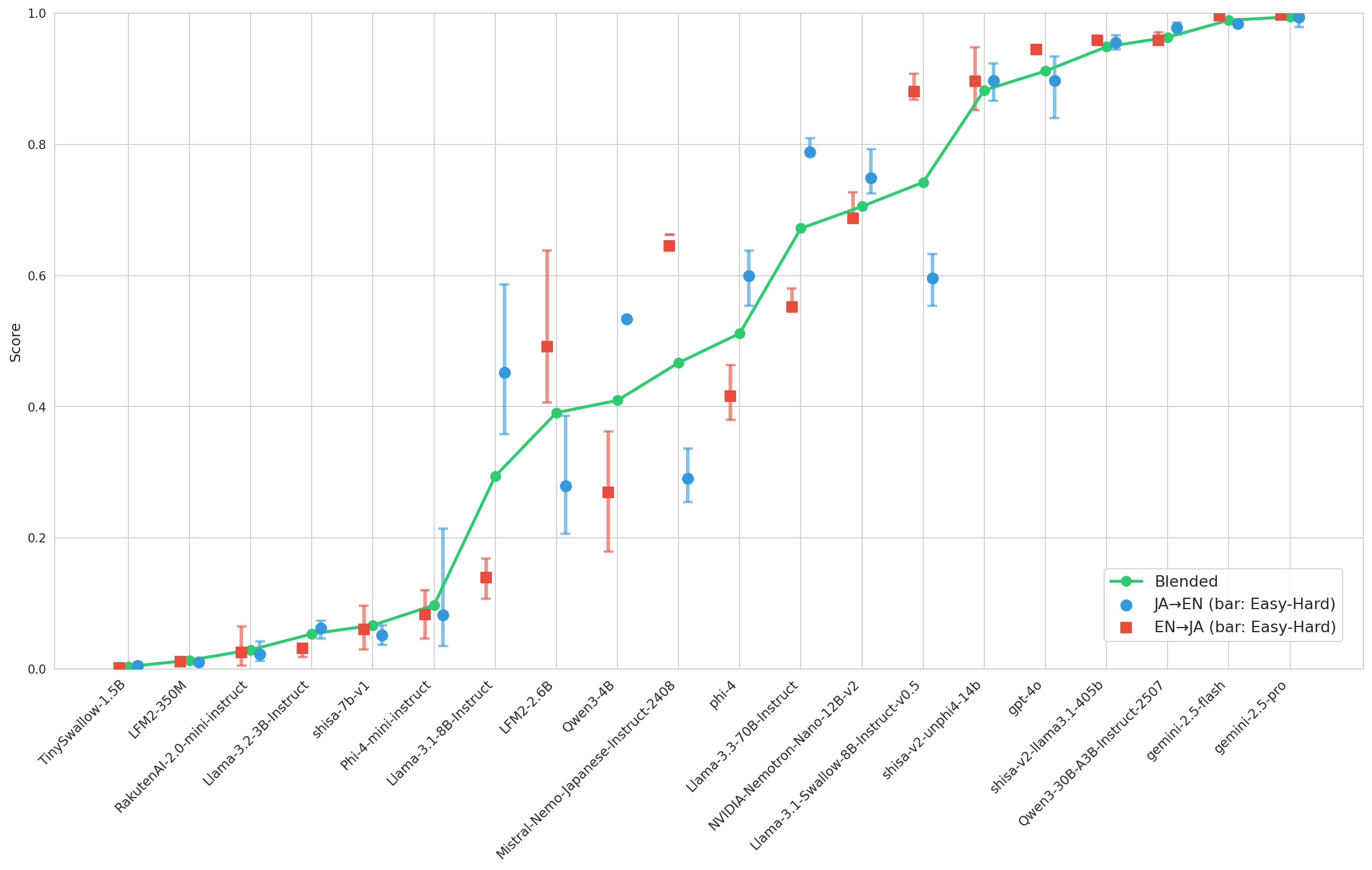}
\caption{JP-TL-Bench Scores by Translation Direction and Difficulty Set. JP-TL-Bench LT scores separated by translation direction (JA$\to$EN in blue, EN$\to$JA in red) with candlestick bars indicating the Easy--Hard range. The blended overall score is shown in green.}
\end{figure}

\subsection{5.2 Comparison with COMET Metrics}\label{comparison-with-comet-metrics}

JP-TL-Bench is still much more resource intensive to run and is not intended to replace learned metrics such as COMET \cite{rei-etal-2020-comet} or evaluation suites such as llm-jp-eval \cite{llmjpeval}. These tools serve complementary purposes:

\begin{itemize}
\tightlist
\item
  \textbf{Use COMET/llm-jp-eval} for broad validation and regression detection on large corpora.
\item
  \textbf{Use JP-TL-Bench} for high-resolution iteration when candidates are already sufficiently performant according to automatic metrics.
\end{itemize}

That being said, COMET-family metrics have significant limitations both in terms of discrimination for top-end performance, and in compressed reporting range. To show this we have run COMET-based evaluations on our Base Set v1.0 translations to allow for direct comparison to our JP-TL-Bench approach:

\begin{itemize}
\tightlist
\item
  \textbf{COMET Ref} (wmt22-comet-da): Reference-based, using gemini-2.5-flash translations as reference
\item
  \textbf{COMET QE} (wmt22-cometkiwi-da): Reference-free quality estimation
\item
  \textbf{X-COMET-XL} \cite{guerreiro2023xcomet}: Reference-based with explainable error spans
\end{itemize}

\begin{figure}[H]
\centering
\includegraphics[width=0.70\textwidth]{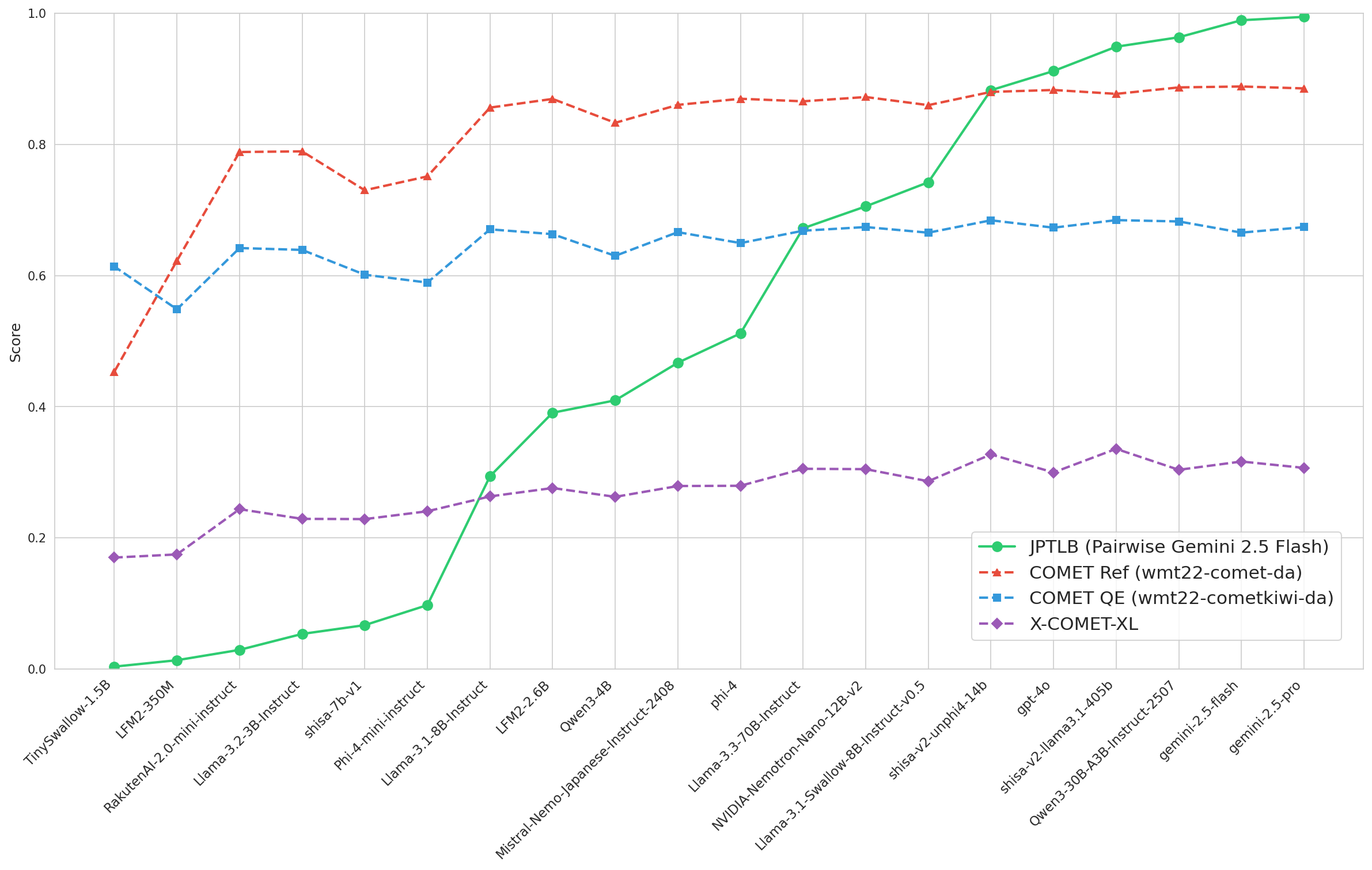}
\caption{Score Progression by Model. Normalized scores (0--1 scale) across all 20 anchor models, sorted by JP-TL-Bench LT score. The critical observation is score compression at the top: while JP-TL-Bench spreads the top 6 models across a meaningful range (LT 8.8--9.9), COMET metrics compress them into narrow bands (~0.87--0.89 for Ref, ~0.66--0.68 for QE). X-COMET-XL shows improved dynamic range compared to the WMT22 models, but still exhibits compression in the 0.30--0.34 range for top performers.}
\end{figure}

\begin{figure}[H]
\centering
\begin{minipage}[t]{0.48\textwidth}
\vspace{0pt}
\centering
\includegraphics[width=\textwidth]{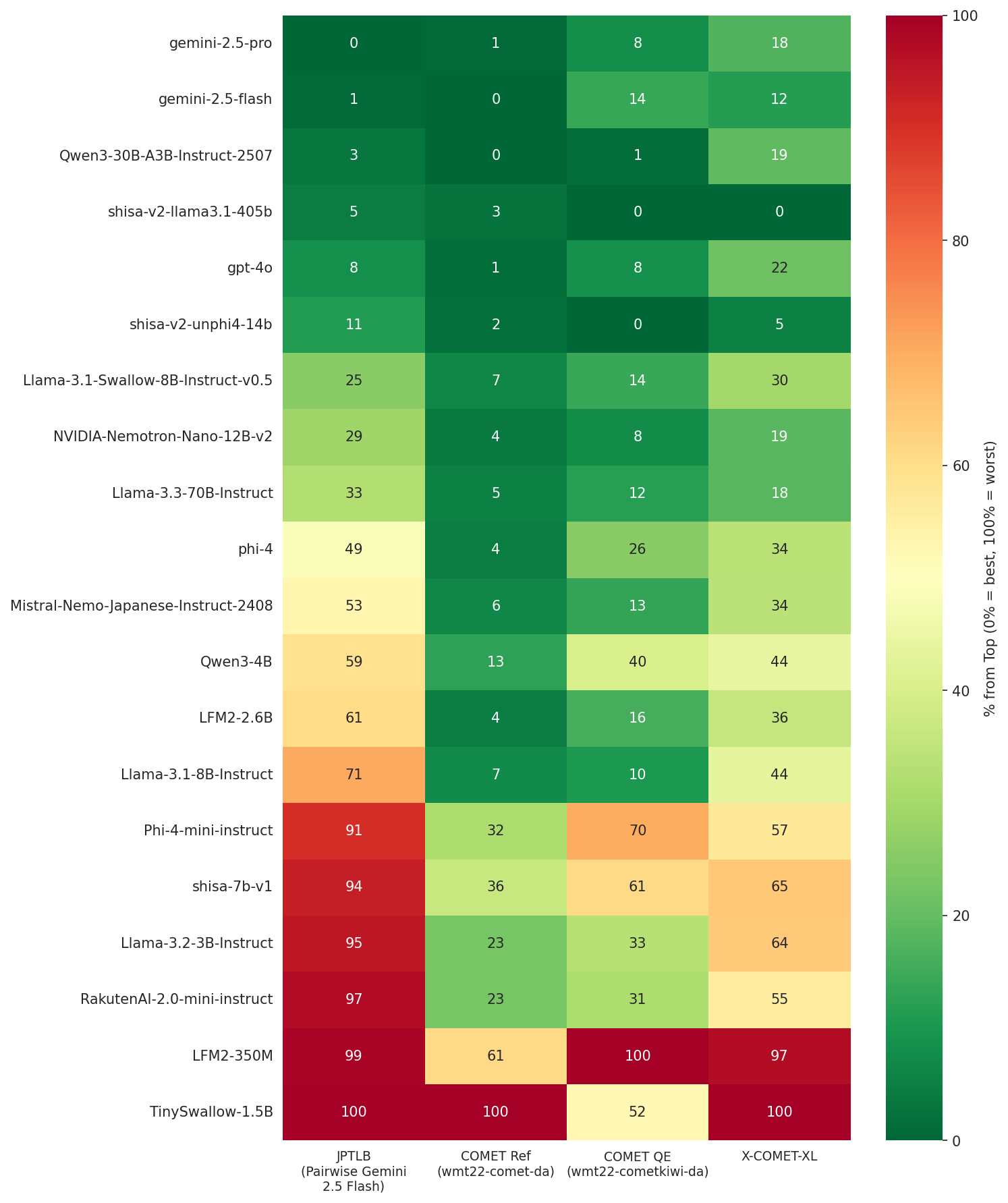}
\caption{Relative Position Heatmap. Each model's position as "\% from top" within each metric (0\% = best, 100\% = worst). COMET metrics cluster many models in the 0--15\% range, while JP-TL-Bench provides more gradual separation across the full range---precisely the property needed for development-time model selection.}
\end{minipage}
\hfill
\begin{minipage}[t]{0.48\textwidth}
\vspace{0pt}
\centering
\includegraphics[width=\textwidth]{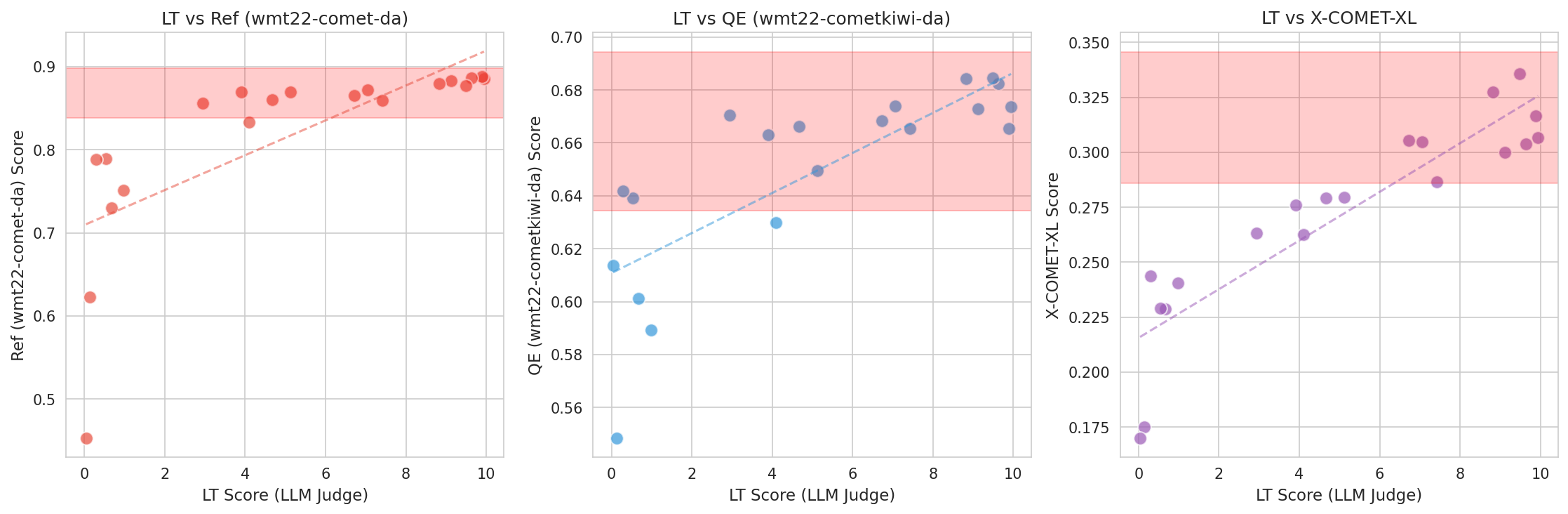}
\caption{Saturation Scatter Plots. The red shaded regions highlight where models with meaningfully different JP-TL-Bench scores cluster at similar COMET values. The positive trend lines confirm that COMET metrics correctly identify quality direction, but the vertical compression limits their utility for discriminating between strong systems.}
\vspace{1em}
\includegraphics[width=\textwidth]{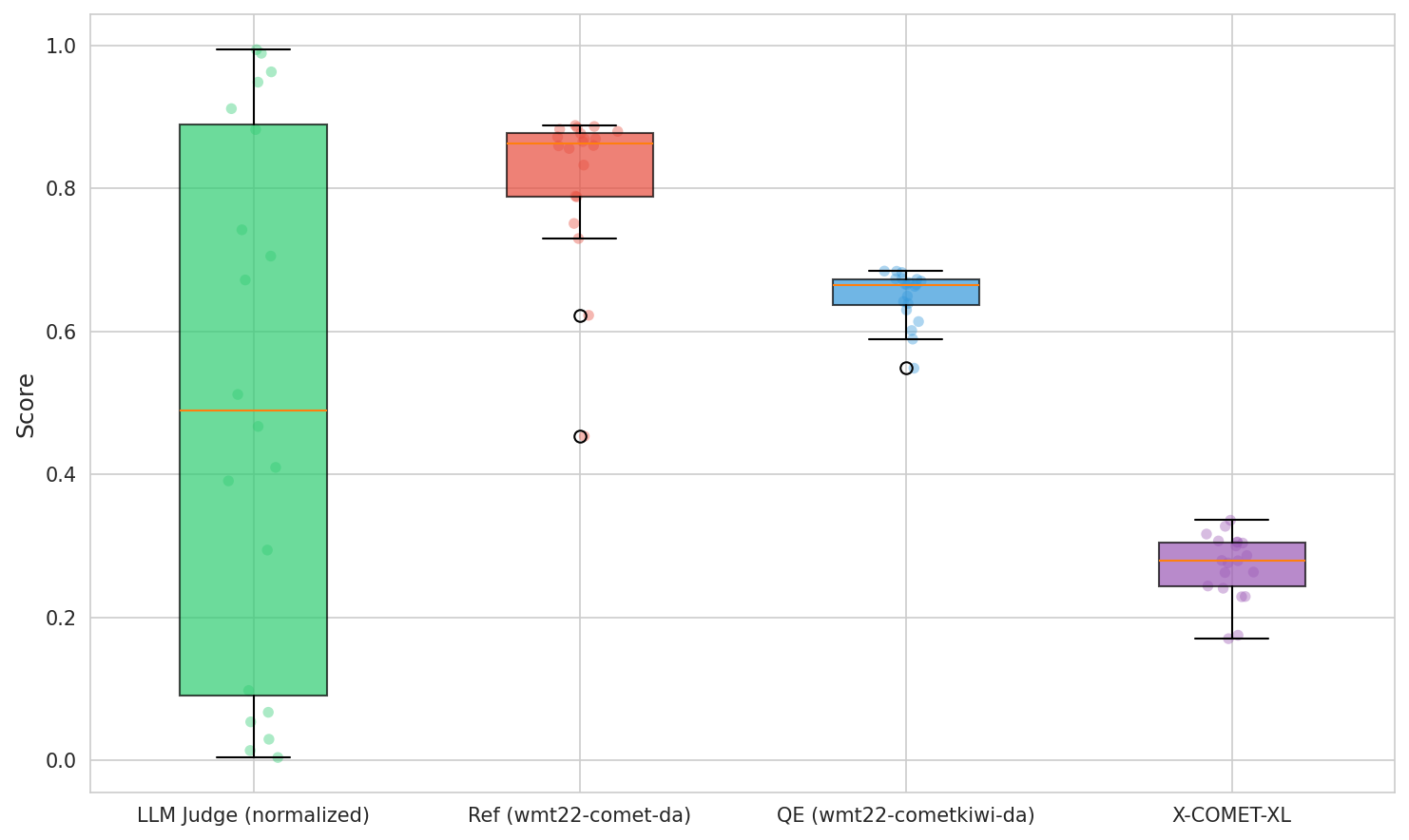}
\caption{Score Distribution Comparison. JP-TL-Bench (normalized) shows the widest interquartile range, providing the most separation between models. All COMET variants produce substantially compressed score ranges.}
\end{minipage}
\end{figure}

\subsection{5.3 Rubric-based judging}\label{rubric-based-judging}

We additionally document rubric-based, reference-aware LLM judging as used in our llm-jp-eval MT/LiquidAI evaluation workflow (Appendix C). Rubric judges provide interpretable scores on a 1--5 scale and can offer better discrimination than COMET-based evaluators \cite{cui2024ultrafeedback}.

\subsubsection{5.3.1 Case study: LFM2-350M-ENJP-MT}\label{case-study-lfm2-350m-enjp-mt}

Liquid AI reports that ``LFM2-350M-ENJP-MT delivers translation quality that is on par with models more than 10 times its size'' \cite{liquidai2025lfm2mt}, and shows a plot of its llm-jp-eval MT score matching not only Gemma 3 4B but also GPT-4o (\emph{slightly} bigger than 10\(\times\) in size). With proper parameter and prompt template tuning, we were able to replicate and confirm these surprising COMET scores, but our qualitative analysis suggests this parity is an artifact of metric compression.

\begin{table}[htbp]
\centering
\small
\caption{COMET scores (llm-jp-eval MT)}
\begin{tabular}{@{}llll@{}}
\toprule
Model & COMET EN$\to$JA & COMET JA$\to$EN & COMET Avg \\
\midrule
shisa-v2-llama3.1-405b & 0.9165 & 0.8936 & 0.9050 \\
GPT-4o & 0.9212 & 0.8973 & 0.9093 \\
LFM2-350M-ENJP-MT & 0.9046 & 0.8731 & 0.8889 \\
gemma-3-4b-it & 0.8926 & 0.8694 & 0.8810 \\
\bottomrule
\end{tabular}
\end{table}

The COMET table illustrates top-end compression: LFM2-350M-ENJP-MT, Gemma 3 4B, GPT-4o, and our own Shisa V2 405B model all score almost exactly the same. Usually when one sees these score plateaus, the common assumption is the benchmark task itself is saturated (i.e., the translation difficulty is too low to distinguish models). However, examining the raw translations reveals clear quality differences. This suggests \textbf{metric saturation} rather than \textbf{task saturation}: an underlying performance gap exists, but COMET lacks the resolution to capture it.

Once rubric-based judging is applied, the scores show meaningful separation that better reflects the observable output quality:

\begin{table}[htbp]
\centering
\scriptsize
\caption{Score distribution by model (rubric-based judge). Percentage columns show share of samples for each 1-5 rating; Useful\% aggregates scores 3 or higher; Perfect\% is the share of 5s.}
\begin{tabular}{@{}lrrrrrrrrrr@{}}
\toprule
Model & Samples & Mean & Median & 1\% & 2\% & 3\% & 4\% & 5\% & Useful\% & Perfect\% \\
\midrule
shisa-v2-llama3.1-405b & 200 & 4.57 & 5.0 & 0.0 & 1.5 & 6.5 & 26.0 & 66.0 & 98.5 & 66.0 \\
GPT-4o & 200 & 4.55 & 5.0 & 0.0 & 0.5 & 7.5 & 28.0 & 64.0 & 99.5 & 64.0 \\
LFM2-350M-ENJP-MT & 200 & 3.96 & 4.0 & 0.0 & 12.5 & 18.5 & 29.5 & 39.5 & 87.5 & 39.5 \\
gemma-3-4b-it & 200 & 3.69 & 4.0 & 0.0 & 13.5 & 25.0 & 40.5 & 21.0 & 86.5 & 21.0 \\
\bottomrule
\end{tabular}
\end{table}

This distribution highlights a critical behavior profile for Small Language Models (SLMs). While the 350M model achieves a remarkable \textbf{87.5\% ``Useful'' rating} (conceptually accurate output), it falters on the \textbf{``Perfect'' metric (39.5\%)}, lagging significantly behind the frontier models. Crucially, this ``adequacy'' does not translate to preference. When subjected to the comparative rigor of JP-TL-Bench, SLMs rank roughly according to capacity expectations (Appendix E). This suggests that while highly-optimized SLMs can satisfy valid-paraphrase metrics (COMET) and broad acceptability checks (Useful\%), they often lack the stylistic nuance and robustness to win head-to-head comparisons against larger models. They are effectively \textbf{``correct but worse''}---a distinction that only pairwise discrimination captures reliably.

Rubric-based judging is useful and there has been much recent work on improving its quality \cite{kim2024prometheus,cui2024ultrafeedback}. However, absolute rubric scores can still compress at the top end and absolute scoring is inherently more sensitive to judge/prompt interactions than relative comparison. In our testing, pairwise comparison yields more reliable and consistent rankings than absolute rubric scoring, which motivated JP-TL-Bench's design (see also Appendix C and D).

\vspace{0.5em}

\section{6. Limitations and Future Work}\label{limitations-and-future-work}
\FloatBarrier

\begin{enumerate}
\def\labelenumi{\arabic{enumi}.}
\tightlist
\item
  \textbf{Judge dependence}: scores depend on the judge model and prompt; different judges can produce different orderings \cite{wang2023large,llmjudge_survey}.
\item
  \textbf{Judge self-preference}: The default judge (gemini-2.5-flash) is also an anchor model, and LLM evaluators have been shown to favor their own outputs and those of related models \cite{panickssery2024selfrecognition}. In our bilingual spot-checks on a subset of items, human reviewers consistently preferred Gemini-family outputs, suggesting the top ranking reflects genuine quality. Nonetheless, users evaluating Gemini-family candidates may prefer an external judge to minimize potential self-preference effects.
\item
  \textbf{Coverage}: 70 items cannot represent all translation domains (long-context, dialogue translation, specialized legal/medical text), but expanding the default test set must be carefully considered as additional samples cause multiplicative growth in pairwise comparisons. The appropriate approach is alternative test sets for each new domain.
\item
  \textbf{Single language pair (today)}: similarly, the protocol generalizes to other languages, but extending to new language pairs while retaining evaluation fidelity requires curating new anchor sets with appropriate win-rate spacing.
\item
  \textbf{Comparability scope}: scores are comparable only under the benchmark contract (base set + judge + code). Cross-version comparisons require explicit calibration.
\end{enumerate}

\vspace{0.5em}

\section{7. Conclusion}\label{conclusion}
\FloatBarrier

JP-TL-Bench provides an anchored, pairwise LLM-judged evaluation protocol for Japanese\(\leftrightarrow\)English translation aimed at development-time discrimination. By freezing a diverse anchor set and aggregating comparisons with a Bradley--Terry model \cite{bradley1952rank}, it produces stable, interpretable scores under a clear reproducibility contract. Unlike COMET-family metrics that compress strong models into indistinguishable bands, JP-TL-Bench maintains meaningful separation across the full quality spectrum. While we focus on Japanese\(\leftrightarrow\)English translation, the anchored pairwise approach generalizes to other language pairs and evaluation domains where existing metrics saturate---the property most needed when iterating on quality for strong LLMs.

\vspace{0.5em}

\clearpage
\section{Appendix A: Prompt Examples}\label{appendix-a-prompt-examples}

\subsection{\texorpdfstring{A.1 Easy Prompt (EN\(\to\)JA)}{A.1 Easy Prompt (EN\textbackslash toJA)}}\label{a.1-easy-prompt-entoja}

\textbf{Contents}: A short sample conversation about a student's first day at school.

\textbf{Sample}: ``Yesterday was my first day at Riverside High, and wow, what a difference from my old school! The campus is huge, with three separate buildings and an amazing courtyard filled with cherry blossom trees\ldots{}''

\subsection{\texorpdfstring{A.2 Hard Prompt (EN\(\to\)JA)}{A.2 Hard Prompt (EN\textbackslash toJA)}}\label{a.2-hard-prompt-entoja}

\textbf{Contents}: A section of dialogue from a popular action video game, \emph{Metal Gear Rising}.

\textbf{Sample}: ``Free will is a myth. Religion is a joke. We are all pawns, controlled by something greater: Memes. The DNA of the soul. They shape our will. They are the culture---they are everything we pass on. Expose someone to anger long enough, they will learn to hate. They become a carrier. Envy, greed, despair\ldots{} All memes. All passed along.''

\subsection{\texorpdfstring{A.3 Hard Prompt (JA\(\to\)EN)}{A.3 Hard Prompt (JA\textbackslash toEN)}}\label{a.3-hard-prompt-jatoen}

\textbf{Contents}: The opening passage to 黒死館殺人事件 (\emph{Kokushikan Satsujin Jiken} / \emph{The Black Death Mansion Murders}), widely considered to be one of the most difficult books ever written in the Japanese language.

\textbf{Sample}: 「聖アレキセイ寺院の殺人事件に法水が解決を公表しなかったので、そろそろ迷宮入りの噂が立ちはじめた十日目のこと、その日から捜査関係の主脳部は、ラザレフ殺害者の追求を放棄しなければならなくなった。と云うのは、四百年の昔から纏綿としていて、臼杵耶蘇会神学林以来の神聖家族と云われる降矢木の館に、突如真黒い風みたいな毒殺者の彷徨が始まったからであった\ldots」

\vspace{0.5em}

\section{Appendix B: JP-TL-Bench Judge Prompt (compare\_prompt.txt)}\label{appendix-b-jp-tl-bench-judge-prompt-compare_prompt.txt}

This is the compare prompt used by JP-TL-Bench's \texttt{translation\_comparer\_any\_model.py}:

\begin{verbatim}
You are an expert in evaluating translations between Japanese and English. Your task is to compare two translations
of the same source text and determine which one better captures the meaning, nuance, and natural flow of the original.

Here are the evaluation criteria:

1. Accuracy: Faithful representation of the source text's meaning
2. Natural Expression: Fluent, idiomatic language in the target language
3. Tone & Register: Appropriate formality level and style for the context
4. Cultural Adaptations: Appropriate handling of cultural references and idioms
5. Technical Precision: Accurate translation of specialized terms and concepts
6. Structural Flow: Natural sentence structure and paragraph organization
7. Consistency: Uniform terminology and style throughout
8. Target Audience Consideration: Appropriateness for intended readers
Note that if a translation is empty, it means a valid answer was not submitted and it loses the comparison by default.

Instructions:
1. Carefully read the source text and both translations
2. Review each translation against the evaluation criteria
3. For each criterion, compare Translation A and Translation B
4. Determine which translation performed better for each criterion
5. Provide specific examples and brief explanations for your decisions

Complete your evaluation with:
<translation_analysis>
[Your detailed analysis of both translations, addressing each criterion with specific examples]
</translation_analysis>

<evaluation_summary>
[Brief summary of key differences and overall assessment]
</evaluation_summary>

<answer>(A or B ONLY, leave all commentary in translation_analysis. This will be machine graded, so only answer
with A or B here.)</answer>

{{formatted_data}}
\end{verbatim}

\vspace{0.5em}

\section{Appendix C: LiquidAI MT Judge (llm-jp-eval hackathon rubric prompt)}\label{appendix-c-liquidai-mt-judge-llm-jp-eval-hackathon-rubric-prompt}

In a workflow built at the LiquidAI Tokyo Hackathon, we extend llm-jp-eval MT to also run an \textbf{absolute}, \textbf{reference-aware}, \textbf{rubric-based} LLM judge to contrast with COMET/BLEU scores. This approach helps investigate high-level metric saturation and is distinct from JP-TL-Bench (which is \textbf{reference-free} and \textbf{pairwise}).

The source code for this analysis is available at: \url{https://github.com/lhl/liquid-ai-hackathon-tokyo/}

\subsection{C.1 Scoring rubric (1--5 + perfect flag)}\label{c.1-scoring-rubric-15-perfect-flag}

The judge assigns a strict 1--5 score:

\begin{itemize}
\tightlist
\item
  1: Completely wrong / untranslated / incomprehensible
\item
  2: Major errors that severely impact comprehension
\item
  3: Adequate (main idea conveyed) but noticeable issues
\item
  4: Good (accurate and natural) with minor imperfections
\item
  5: Excellent (professional quality)
\end{itemize}

Additionally, it marks a binary ``perfect'' flag (\texttt{correct=1}) for native-quality outputs with no errors.

\subsection{C.2 Judge prompt template (judge.j2)}\label{c.2-judge-prompt-template-judge.j2}

This is the template used by our judge (abridged only by formatting; content preserved):

\begin{verbatim}
You are a bilingual expert evaluating machine translation quality between English and Japanese.

**Source text**: {{ source_text }}

**Translation**: {{ translated_text }}

**Reference translation** (gold standard): {{ reference_text }}

Evaluate the translation strictly on:
- **Accuracy**: Does it preserve all facts, entities, and meaning from the source?
- **Fluency**: Is it natural, grammatical, and idiomatic in the target language?
- **Completeness**: Is everything translated without omissions or hallucinations?

Use this 1-5 scoring rubric:
- **1** -- Completely wrong, untranslated, or incomprehensible
- **2** -- Major errors in meaning, facts, or grammar that severely impact comprehension
- **3** -- Adequate translation with the main idea conveyed, but noticeable issues in accuracy or fluency
- **4** -- Good translation that is accurate and natural, with only minor imperfections
- **5** -- Excellent translation that is precise, fluent, and indistinguishable from a professional human translation

Additionally, mark **<correct>1</correct>** if the translation is perfect and would be acceptable as-is in a
professional setting (native-quality with no errors). Otherwise use **<correct>0</correct>**.

Respond with a brief justification followed by your scores:
<justification>...</justification><score>N</score><correct>0|1</correct>
\end{verbatim}

\vspace{0.5em}

\section{Appendix D: Pairwise Comparisons vs Numeric Scoring}\label{appendix-d-pairwise-comparisons-vs-numeric-scoring}

Appendix C described a rubric-based approach; here we explain more of why JP-TL-Bench ultimately chose to use pairwise comparison.

\textbf{Source (Japanese):}

\begin{quote}
華やかでおしゃれな街が多く、自然にも恵まれた兵庫県。県内には41の市町があり、県庁所在地の神戸市は中央区や東灘区など9つの行政区に分かれています。それぞれの市町や区には、その土地ならではの魅力がありますよね
\end{quote}

\textbf{Mistral 7B:}

\begin{quote}
The prefecture of Hyogo is known for its vibrant, fashionable cities that are abundantly blessed by nature. There are 41 cities, towns within the prefecture, with Kobe City as its capital being divided into nine administrative districts such as Chuou Ward and Higashinada Ward. Each city, town, or district has unique charms specific to their land, don't they?
\end{quote}

\textbf{Shisa Chotto:}

\begin{quote}
Hyogo Prefecture boasts many vibrant and stylish cities as well as abundant natural beauty. The prefecture consists of 41 cities and towns, with its capital, Kobe City, divided into nine administrative wards such as Chuo Ward and Higashinada Ward. Each city, town, and ward has its own unique charm, doesn't it?
\end{quote}

Above is a comparison between two LLM generated translations of a simple Japanese passage. Both are accurate and contain essentially the same content. However, Mistral 7B contains minor grammatical errors (a comma instead of an ``and'' after cities), slightly odd phrasings like ``abundantly blessed by nature'', and overly literal phrasings such as ``unique charms specific to their land''. Representing these issues with a numeric score is tricky: how does one score a single comma splice in a long text passage? How does someone decide if a phrase like `specific to their land' is awkward enough to penalize, and if so, how much should the penalty be? 3 points out of 100? 5 points?

This is a single sample, but it illustrates the larger principle: given the infinite variety of possible translations for a given piece of text, pairwise comparison yields more consistent judgments. Evaluators readily agree Chotto's translation is superior, even if they would assign different numeric scores. This consistency advantage explains why pairwise comparison produces more reliable rankings than absolute scoring (Section 5.3). By doing hundreds of comparisons like this, we can get a strong picture of a given model's relative strength at complex, real-world translation.

\vspace{0.5em}

\section{Appendix E: Reproducibility Checklist (JP-TL-Bench)}\label{appendix-e-reproducibility-checklist-jp-tl-bench}

When reporting JP-TL-Bench scores:

\begin{itemize}
\tightlist
\item[$\square$]
  Base Set snapshot version (e.g., \texttt{baseset/v1.0})
\item[$\square$]
  Judge model identifier and provider/version (if applicable)
\item[$\square$]
  Judge prompt version (hash/path) and decoding settings (temperature, etc.)
\item[$\square$]
  Candidate model identifier and decoding settings
\item[$\square$]
  Any filtering/backfill applied to missing judgments
\item[$\square$]
  Links to raw comparison logs (or hashes) when possible
\end{itemize}

\vspace{0.5em}

\section{Appendix F: Full Model Scores}\label{appendix-f-full-model-scores}

For reference, we include JP-TL-Bench v1.0 scores for a selection of models evaluated during development:

\subsection{\texorpdfstring{EN\(\to\)JA: English to Japanese Translation}{EN\textbackslash toJA: English to Japanese Translation}}\label{entoja-english-to-japanese-translation}

\footnotesize
\begin{longtable}{@{}lrrrr@{}}
\toprule
Model & Easy LT & Hard LT & Overall LT & Win Rate \\
\midrule
\endfirsthead
\toprule
Model & Easy LT & Hard LT & Overall LT & Win Rate \\
\midrule
\endhead
\bottomrule
\endfoot
\texttt{google/gemini-3-flash-preview} & 9.98 & 10.00 & 9.99 & 97.3\% \\
\texttt{google/gemini-3-pro-preview} & 9.99 & 9.99 & 9.99 & 97.5\% \\
\texttt{google/gemini-2.5-pro} & 9.95 & 9.99 & 9.97 & 96.6\% \\
\texttt{google/gemini-2.5-flash} & 9.97 & 9.98 & 9.96 & 95.4\% \\
\texttt{deepseek-ai/DeepSeek-V3.1-Terminus} & 9.96 & 9.94 & 9.94 & 91.6\% \\
\texttt{openai/gpt-oss-120b} & 9.85 & 9.91 & 9.86 & 87.2\% \\
\texttt{moonshotai/Kimi-K2-Instruct-0905} & 9.86 & 9.90 & 9.86 & 87.5\% \\
\texttt{Qwen/Qwen3-235B-A22B-Instruct-2507} & 9.91 & 9.84 & 9.85 & 87.8\% \\
\texttt{google/gemma-3-27b-it} & 9.79 & 9.66 & 9.69 & 82.2\% \\
\texttt{shisa-ai/shisa-v2-llama3.1-405b} & 9.74 & 9.69 & 9.67 & 80.7\% \\
\texttt{shisa-ai/chotto} & 9.77 & 9.60 & 9.65 & 81.1\% \\
\texttt{shisa-ai/shisa-v2.1-unphi4-14b} & 9.59 & 9.74 & 9.63 & 81.6\% \\
\texttt{shisa-ai/shisa-v2.1-llama3.3-70b} & 9.50 & 9.79 & 9.63 & 80.4\% \\
gpt-4o-2024-08-06 & 9.61 & 9.68 & 9.61 & 80.1\% \\
\texttt{Qwen/Qwen3-30B-A3B-Instruct-2507} & 9.48 & 9.72 & 9.56 & 79.9\% \\
\texttt{mistralai/Ministral-3-14B-Instruct-2512} & 9.43 & 9.71 & 9.55 & 78.9\% \\
\texttt{inclusionAI/Ling-1T-FP8} & 9.68 & 9.37 & 9.48 & 77.9\% \\
\texttt{shisa-ai/shisa-v2-llama3.3-70b} & 9.08 & 9.64 & 9.36 & 76.0\% \\
\texttt{shisa-ai/shisa-v2-unphi4-14b} & 9.65 & 9.39 & 9.27 & 76.5\% \\
\texttt{shisa-ai/shisa-v2.1-qwen3-8b} & 9.23 & 9.34 & 9.21 & 75.1\% \\
\texttt{abeja/ABEJA-Qwen2.5-32b-Japanese-v1.0} & 9.24 & 9.08 & 9.07 & 72.2\% \\
\texttt{elyza/ELYZA-Shortcut-1.0-Qwen-32B} & 9.01 & 9.26 & 9.07 & 71.9\% \\
\texttt{tokyotech-llm/Llama-3.3-Swallow-70B-Instruct-v0.4} & 9.26 & 9.03 & 9.06 & 71.7\% \\
\texttt{stockmark/Stockmark-2-100B-Instruct} & 9.12 & 8.88 & 8.90 & 69.9\% \\
\texttt{elyza/ELYZA-Thinking-1.0-Qwen-32B} & 8.86 & 9.06 & 8.88 & 70.1\% \\
\texttt{tokyotech-llm/Llama-3.1-Swallow-8B-Instruct-v0.5} & 8.91 & 9.00 & 8.87 & 70.5\% \\
\texttt{unsloth/phi-4} & 8.87 & 8.75 & 8.70 & 68.4\% \\
\texttt{shisa-ai/shisa-v2.1-llama3.2-3b} & 8.48 & 9.01 & 8.68 & 68.4\% \\
\texttt{flux-inc/Flux-Japanese-Qwen2.5-32B-Instruct-V1.0} & 8.35 & 8.76 & 8.46 & 66.0\% \\
\texttt{nvidia/NVIDIA-Nemotron-3-Nano-30B-A3B-BF16} & 7.16 & 7.32 & 7.10 & 58.5\% \\
\texttt{meta-llama/Llama-3.1-405B-Instruct} & 5.15 & 8.45 & 7.05 & 58.3\% \\
\texttt{shisa-ai/shisa-v2-llama3.1-8b} & 7.41 & 6.77 & 6.93 & 57.8\% \\
\texttt{nvidia/NVIDIA-Nemotron-Nano-12B-v2} & 7.27 & 6.85 & 6.87 & 58.2\% \\
\texttt{inclusionAI/Ling-flash-2.0} & 6.11 & 7.40 & 6.72 & 55.8\% \\
\texttt{Qwen/Qwen3-8B} & 6.85 & 6.86 & 6.72 & 57.9\% \\
\texttt{cyberagent/Mistral-Nemo-Japanese-Instruct-2408} & 6.62 & 6.62 & 6.45 & 56.2\% \\
\texttt{elyza/ELYZA-Shortcut-1.0-Qwen-7B} & 6.62 & 6.38 & 6.32 & 54.3\% \\
\texttt{shisa-ai/shisa-v2.1-lfm2-1.2b} & 6.66 & 5.30 & 5.75 & 51.6\% \\
\texttt{meta-llama/Llama-3.3-70B-Instruct} & 5.20 & 6.17 & 5.59 & 52.4\% \\
\texttt{LiquidAI/LFM2-8B-A1B} & 4.79 & 6.40 & 5.57 & 50.7\% \\
\texttt{LiquidAI/LFM2-2.6B} & 6.24 & 4.44 & 5.07 & 49.5\% \\
\texttt{openai/gpt-oss-20b} & 8.35 & 2.41 & 4.92 & 48.7\% \\
\texttt{sbintuitions/sarashina2.2-3b-instruct-v0.1} & 4.46 & 5.41 & 4.86 & 48.0\% \\
\texttt{microsoft/phi-4} & 3.80 & 4.64 & 4.16 & 46.1\% \\
\texttt{mistralai/Ministral-3-3B-Instruct-2512} & 3.46 & 4.52 & 3.95 & 44.2\% \\
\texttt{meta-llama/Llama-4-Scout-17B-16E} & 1.92 & 5.41 & 3.67 & 43.2\% \\
\texttt{baidu/ERNIE-4.5-21B-A3B-PT} & 3.11 & 3.71 & 3.42 & 41.9\% \\
\texttt{Qwen/Qwen3-4B} & 1.79 & 3.62 & 2.70 & 39.3\% \\
\texttt{meta-llama/Llama-3.2-1B-Instruct} & 1.24 & 2.44 & 1.88 & 30.7\% \\
\texttt{LiquidAI/LFM2-1.2B} & 2.07 & 1.27 & 1.63 & 33.2\% \\
\texttt{baidu/ERNIE-4.5-VL-28B-A3B-PT} & 1.36 & 1.46 & 1.47 & 30.6\% \\
\texttt{meta-llama/Llama-3.1-8B-Instruct} & 1.01 & 1.66 & 1.37 & 31.2\% \\
\texttt{microsoft/Phi-4-mini-instruct} & 0.77 & 1.07 & 0.97 & 28.0\% \\
\texttt{Nanbeige/Nanbeige4-3B-Thinking-2511} & 0.58 & 1.04 & 0.87 & 25.0\% \\
\texttt{augmxnt/shisa-7b-v1} & 0.91 & 0.46 & 0.70 & 24.7\% \\
\texttt{augmxnt/shisa-gamma-7b-v1} & 0.76 & 0.32 & 0.54 & 20.7\% \\
\texttt{openbmb/MiniCPM4.1-8B} & 0.35 & 0.62 & 0.54 & 21.0\% \\
\texttt{microsoft/Phi-4-multimodal-instruct} & 1.09 & 0.17 & 0.52 & 20.8\% \\
\texttt{meta-llama/Llama-3.2-3B-Instruct} & 0.19 & 0.35 & 0.32 & 17.2\% \\
\texttt{allenai/Olmo-3-7B-Instruct} & 0.15 & 0.30 & 0.28 & 16.5\% \\
\texttt{Rakuten/RakutenAI-2.0-mini-instruct} & 0.65 & 0.05 & 0.25 & 15.3\% \\
\texttt{LiquidAI/LFM2-350M} & 0.08 & 0.08 & 0.12 & 11.8\% \\
\texttt{mistralai/Mistral-7B-Instruct-v0.1} & 0.08 & 0.06 & 0.09 & 8.9\% \\
\texttt{SakanaAI/TinySwallow-1.5B} & 0.01 & 0.01 & 0.01 & 1.4\% \\
\end{longtable}
\normalsize

\subsection{\texorpdfstring{JA\(\to\)EN: Japanese to English Translation}{JA\textbackslash toEN: Japanese to English Translation}}\label{jatoen-japanese-to-english-translation}

\footnotesize
\begin{longtable}{@{}lrrrr@{}}
\toprule
Model & Easy LT & Hard LT & Overall LT & Win Rate \\
\midrule
\endfirsthead
\toprule
Model & Easy LT & Hard LT & Overall LT & Win Rate \\
\midrule
\endhead
\bottomrule
\endfoot
\texttt{google/gemini-3-flash-preview} & 9.94 & 9.99 & 9.98 & 97.4\% \\
\texttt{google/gemini-3-pro-preview} & 9.88 & 9.98 & 9.94 & 94.1\% \\
\texttt{google/gemini-2.5-pro} & 9.79 & 9.99 & 9.94 & 95.7\% \\
\texttt{Qwen/Qwen3-235B-A22B-Instruct-2507} & 9.83 & 9.97 & 9.93 & 93.2\% \\
\texttt{openai/gpt-oss-120b} & 9.89 & 9.95 & 9.92 & 92.6\% \\
\texttt{deepseek-ai/DeepSeek-V3.1-Terminus} & 9.85 & 9.93 & 9.89 & 90.9\% \\
\texttt{inclusionAI/Ling-1T-FP8} & 9.93 & 9.89 & 9.89 & 91.0\% \\
\texttt{mistralai/Ministral-3-14B-Instruct-2512} & 9.90 & 9.89 & 9.88 & 90.7\% \\
\texttt{moonshotai/Kimi-K2-Instruct-0905} & 9.79 & 9.92 & 9.86 & 89.3\% \\
\texttt{google/gemini-2.5-flash} & 9.83 & 9.88 & 9.84 & 90.4\% \\
\texttt{shisa-ai/chotto} & 9.96 & 9.70 & 9.79 & 87.4\% \\
\texttt{Qwen/Qwen3-30B-A3B-Instruct-2507} & 9.70 & 9.84 & 9.77 & 86.5\% \\
\texttt{shisa-ai/shisa-v2-llama3.3-70b} & 9.74 & 9.83 & 9.77 & 85.9\% \\
\texttt{shisa-ai/shisa-v2-llama3.1-405b} & 9.65 & 9.83 & 9.74 & 83.5\% \\
\texttt{google/gemma-3-27b-it} & 9.54 & 9.77 & 9.66 & 82.7\% \\
\texttt{shisa-ai/shisa-v2.1-llama3.3-70b} & 9.43 & 9.80 & 9.66 & 82.5\% \\
gpt-4o-2024-08-06 & 9.59 & 9.74 & 9.65 & 82.4\% \\
\texttt{shisa-ai/shisa-v2.1-unphi4-14b} & 9.55 & 9.28 & 9.36 & 78.5\% \\
\texttt{shisa-ai/shisa-v2-unphi4-14b} & 9.44 & 9.20 & 9.17 & 76.5\% \\
\texttt{shisa-ai/shisa-v2.1-qwen3-8b} & 8.77 & 9.26 & 9.02 & 73.7\% \\
\texttt{inclusionAI/Ling-flash-2.0} & 8.85 & 9.19 & 9.00 & 72.3\% \\
\texttt{abeja/ABEJA-Qwen2.5-32b-Japanese-v1.0} & 8.36 & 9.08 & 8.74 & 69.3\% \\
\texttt{unsloth/phi-4} & 8.61 & 8.90 & 8.72 & 69.2\% \\
\texttt{nvidia/NVIDIA-Nemotron-3-Nano-30B-A3B-BF16} & 9.47 & 8.00 & 8.69 & 70.1\% \\
\texttt{tokyotech-llm/Llama-3.3-Swallow-70B-Instruct-v0.4} & 8.15 & 8.89 & 8.57 & 68.1\% \\
\texttt{elyza/ELYZA-Shortcut-1.0-Qwen-32B} & 8.04 & 8.94 & 8.53 & 67.1\% \\
\texttt{meta-llama/Llama-3.1-405B-Instruct} & 9.05 & 8.12 & 8.49 & 67.0\% \\
\texttt{elyza/ELYZA-Thinking-1.0-Qwen-32B} & 7.73 & 8.80 & 8.34 & 65.6\% \\
\texttt{shisa-ai/shisa-v2-llama3.1-8b} & 8.10 & 8.56 & 8.30 & 65.2\% \\
\texttt{meta-llama/Llama-3.3-70B-Instruct} & 8.46 & 8.09 & 8.19 & 65.5\% \\
\texttt{stockmark/Stockmark-2-100B-Instruct} & 8.57 & 7.97 & 8.18 & 64.7\% \\
\texttt{openai/gpt-oss-20b} & 8.47 & 7.96 & 8.12 & 65.3\% \\
\texttt{Qwen/Qwen3-8B} & 6.95 & 8.63 & 7.94 & 64.6\% \\
\texttt{flux-inc/Flux-Japanese-Qwen2.5-32B-Instruct-V1.0} & 7.32 & 8.09 & 7.75 & 62.2\% \\
\texttt{nvidia/NVIDIA-Nemotron-Nano-12B-v2} & 7.93 & 7.25 & 7.49 & 61.7\% \\
\texttt{mistralai/Ministral-3-3B-Instruct-2512} & 7.97 & 6.88 & 7.29 & 58.2\% \\
\texttt{shisa-ai/shisa-v2.1-llama3.2-3b} & 8.50 & 5.86 & 7.09 & 57.3\% \\
\texttt{baidu/ERNIE-4.5-21B-A3B-PT} & 6.28 & 7.69 & 7.07 & 57.2\% \\
\texttt{tokyotech-llm/Llama-3.1-Swallow-8B-Instruct-v0.5} & 6.25 & 7.10 & 6.69 & 56.5\% \\
\texttt{microsoft/phi-4} & 5.54 & 6.38 & 6.00 & 53.6\% \\
\texttt{baidu/ERNIE-4.5-VL-28B-A3B-PT} & 6.12 & 5.87 & 5.94 & 51.7\% \\
\texttt{Qwen/Qwen3-4B} & 5.38 & 5.36 & 5.34 & 50.4\% \\
\texttt{elyza/ELYZA-Shortcut-1.0-Qwen-7B} & 4.85 & 5.21 & 5.01 & 46.9\% \\
\texttt{sbintuitions/sarashina2.2-3b-instruct-v0.1} & 4.66 & 4.93 & 4.79 & 46.0\% \\
\texttt{meta-llama/Llama-3.1-8B-Instruct} & 6.07 & 3.74 & 4.70 & 47.1\% \\
\texttt{meta-llama/Llama-4-Scout-17B-16E} & 4.82 & 3.17 & 3.84 & 41.9\% \\
\texttt{LiquidAI/LFM2-8B-A1B} & 4.17 & 2.80 & 3.37 & 39.3\% \\
\texttt{LiquidAI/LFM2-2.6B} & 4.06 & 2.38 & 3.06 & 38.8\% \\
\texttt{cyberagent/Mistral-Nemo-Japanese-Instruct-2408} & 3.37 & 2.55 & 2.91 & 38.8\% \\
\texttt{Nanbeige/Nanbeige4-3B-Thinking-2511} & 4.29 & 1.93 & 2.87 & 36.5\% \\
\texttt{openbmb/MiniCPM4.1-8B} & 3.86 & 1.79 & 2.57 & 35.5\% \\
\texttt{microsoft/Phi-4-multimodal-instruct} & 2.72 & 1.18 & 1.75 & 30.8\% \\
\texttt{allenai/Olmo-3-7B-Instruct} & 1.92 & 1.25 & 1.57 & 29.6\% \\
\texttt{shisa-ai/shisa-v2.1-lfm2-1.2b} & 2.89 & 0.68 & 1.37 & 28.0\% \\
\texttt{microsoft/Phi-4-mini-instruct} & 2.22 & 0.82 & 1.31 & 29.3\% \\
\texttt{meta-llama/Llama-3.2-3B-Instruct} & 0.94 & 0.61 & 0.80 & 24.9\% \\
\texttt{augmxnt/shisa-7b-v1} & 0.82 & 0.45 & 0.63 & 22.6\% \\
\texttt{augmxnt/shisa-gamma-7b-v1} & 0.99 & 0.35 & 0.61 & 21.2\% \\
\texttt{LiquidAI/LFM2-1.2B} & 0.83 & 0.27 & 0.48 & 19.6\% \\
\texttt{mistralai/Mistral-7B-Instruct-v0.1} & 0.59 & 0.15 & 0.30 & 14.8\% \\
\texttt{Rakuten/RakutenAI-2.0-mini-instruct} & 0.42 & 0.12 & 0.23 & 13.1\% \\
\texttt{meta-llama/Llama-3.2-1B-Instruct} & 0.13 & 0.11 & 0.14 & 9.9\% \\
\texttt{LiquidAI/LFM2-350M} & 0.08 & 0.08 & 0.12 & 11.8\% \\
\texttt{SakanaAI/TinySwallow-1.5B} & 0.03 & 0.04 & 0.05 & 4.0\% \\
\end{longtable}
\normalsize

\bibliographystyle{unsrt}
\bibliography{references}

\end{CJK*}
\end{document}